\documentclass{article}
\pdfoutput=1

\usepackage{xspace}
\usepackage{xr-hyper}
\usepackage{float} 
\usepackage{dblfloatfix}
\usepackage{url}
\usepackage{enumitem}
\usepackage{multirow}
\usepackage{caption}
\usepackage{wrapfig}
\usepackage{float}
\usepackage{amssymb}
\usepackage{amsmath}
\usepackage{comment}
\usepackage{blindtext}
\usepackage{subfiles}
\usepackage[usenames,dvipsnames]{xcolor}
\newcommand{\rl}{RL\xspace}
\newcommand{\mujoco}{{\sc mujoco}\xspace}
\newcommand{\cvae}{{\sc c-vae}\xspace}

\usepackage[accepted]{icml2020}

\PassOptionsToPackage{numbers, compress}{natbib}
\PassOptionsToPackage{hyphens}{url}\usepackage{hyperref}
\definecolor{myblue}{rgb}{0.03,0.25,0.47}
\hypersetup{
    colorlinks=true,
    linkcolor=myblue,
    citecolor=myblue,
    filecolor=black,
    urlcolor=myblue,
}

\usepackage{import}
\usepackage{microtype}
\usepackage{graphicx}
\usepackage{subfigure}
\makeatletter
\renewcommand{\thesubfigure}{\alph{subfigure}}
\renewcommand{\@thesubfigure}{(\thesubfigure)\hskip\subfiglabelskip}
\makeatother
\usepackage{wrapfig}
\usepackage{booktabs} 

\usepackage[utf8]{inputenc} 
\usepackage[T1]{fontenc}    
\usepackage{url}            
\usepackage{booktabs}       
\usepackage{amsfonts}       
\usepackage{nicefrac}       
\usepackage{microtype}      

\newcommand{\NL}{\textsc{nl}\xspace}

\definecolor{myred}{rgb}{0.8,0,0}

\icmltitlerunning{Language-Conditioned Goal Generation: a New Approach to Language Grounding in RL}

\definecolor{myred}{rgb}{0.8,0,0}
\definecolor{mypurple}{rgb}{0.6,0.22,0.8}

\definecolor{mygreen}{rgb}{0,0.6,0}
\definecolor{myblue}{rgb}{0,0,0.7}
\definecolor{myindigo}{rgb}{0.4,0,0.7}

\begin{document}

\setlength{\abovedisplayskip}{1pt}
\setlength{\belowdisplayskip}{1pt}

\twocolumn[
\icmltitle{Language-Conditioned Goal Generation: \\ a New Approach to Language Grounding in RL}

\icmlsetsymbol{equal}{*}

\begin{icmlauthorlist}
\icmlauthor{C\'edric Colas*}{in}
\icmlauthor{Ahmed Akakzia*}{up}
\icmlauthor{Pierre-Yves Oudeyer}{in}
\icmlauthor{Mohamed Chetouani}{up}
\icmlauthor{Olivier Sigaud}{up}

\end{icmlauthorlist}
\icmlaffiliation{in}{Flowers Team, Inria Bordeaux, FR.}
\icmlaffiliation{up}{Université Paris-Sorbonne, Paris, FR.}
\icmlcorrespondingauthor{C\'edric Colas}{cedric.colas@inria.fr}
\icmlkeywords{generalization, reinforcement learning, autonomous learning,  natural language grounding, symbol grounding}
\vskip 0.3in
]

\printAffiliationsAndNotice{\icmlEqualContribution} 

\begin{abstract}
In the real world, linguistic agents are also embodied agents: they perceive and act in the physical world. The notion of \textit{Language Grounding} questions the interactions between language and embodiment: how do learning agents connect or \textit{ground} linguistic representations to the physical world ? This question has recently been approached by the Reinforcement Learning community under the framework of \textit{instruction-following} agents. In these agents, behavioral policies or reward functions are conditioned on the embedding of an instruction expressed in natural language. This paper proposes another approach: using language to condition goal generators. Given any goal-conditioned policy, one could train a language-conditioned goal generator to generate language-agnostic goals for the agent. This method allows to decouple sensorimotor learning from language acquisition and enable agents to demonstrate a diversity of behaviors for any given instruction. We propose a particular instantiation of this approach and demonstrate its benefits. 
\end{abstract}


\section{Introduction}
\label{sec:introduction}
\textit{Language Grounding} describes the idea that language acquisition is strongly shaped by one's experience of the physical world. This idea emerged in Cognitive Science \cite{harnad1990symbol,Glenberg2002,Zwaan05} and quickly inspired Artificial Intelligence approaches in natural language processing \cite{roy2005connecting}, human-machine interactions \cite{Dominey2005, Madden2010} and more recently deep Reinforcement Learning (deep \rl) \cite{luketina2019survey}.


In the \rl community, this has taken the form of \textit{language-conditioned} agents \cite{hermann2017grounded,chan2019actrce,bahdanau2018learning,cideron2019self,jiang2019language,zhong2019rtfm,waytowich2019grounding,colas2020language}. Language can be used to build representations \cite{waytowich2019grounding} or to characterize the dynamics of the environment \cite{zhong2019rtfm}. However, it is mostly used to represent instructions or goals \cite{hermann2017grounded,chan2019actrce,bahdanau2018learning,cideron2019self,jiang2019language,colas2020language}. In these approaches, natural language (\NL) sentences are embedded through recurrent networks and merged with the agent's state to form the input of the policy or reward function. The language encoder is then trained jointly with either the former \cite{hermann2017grounded,chan2019actrce,jiang2019language,cideron2019self,hill2019emergent} or the latter \cite{bahdanau2018learning,colas2020language}. These approach demonstrate benefits over traditional goal-conditioned methods:
\begin{enumerate}[leftmargin=0.6cm, nolistsep]
    \item 
    \textit{Targeting abstract goals.} Language can express abstract goals characterized by sets of properties the scene should verify (e.g. \textit{block A} above \textit{block B}). This contrasts with previous goal-as-state approaches, where goals are specific states of the agent (e.g. target pixel image, target block positions). 
    \item 
    \textit{Systematic generalization.} Previous language-conditioned approaches demonstrate strong generalization capabilities including generalizations to new combinations of action verbs and object attributes (colors, shapes, categories etc.) \cite{hermann2017grounded,bahdanau2018learning,hill2019emergent,colas2020language}. 
\end{enumerate}
However, this comes with drawbacks:
\begin{enumerate}[leftmargin=0.6cm, nolistsep]
    \item
    \textit{Language becomes a pre-requisite for sensorimotor learning.} Pre-verbal infants are known to demonstrate goal-oriented behavior \cite{wood1976role}. Language-conditioned agents require language inputs to act in the world and, thus, cannot account for this decoupling.
    \item 
    \textit{Lack of behavioral diversity.} Because policies are directly conditioned on language embeddings, an instruction in a given context only generates a low diversity of behaviors: a unique behavior for a deterministic policy and minor noise-induced behavioral variations for stochastic policies. 
\end{enumerate}


\begin{figure}[h!]
    \centering
    \includegraphics[width=0.95\linewidth]{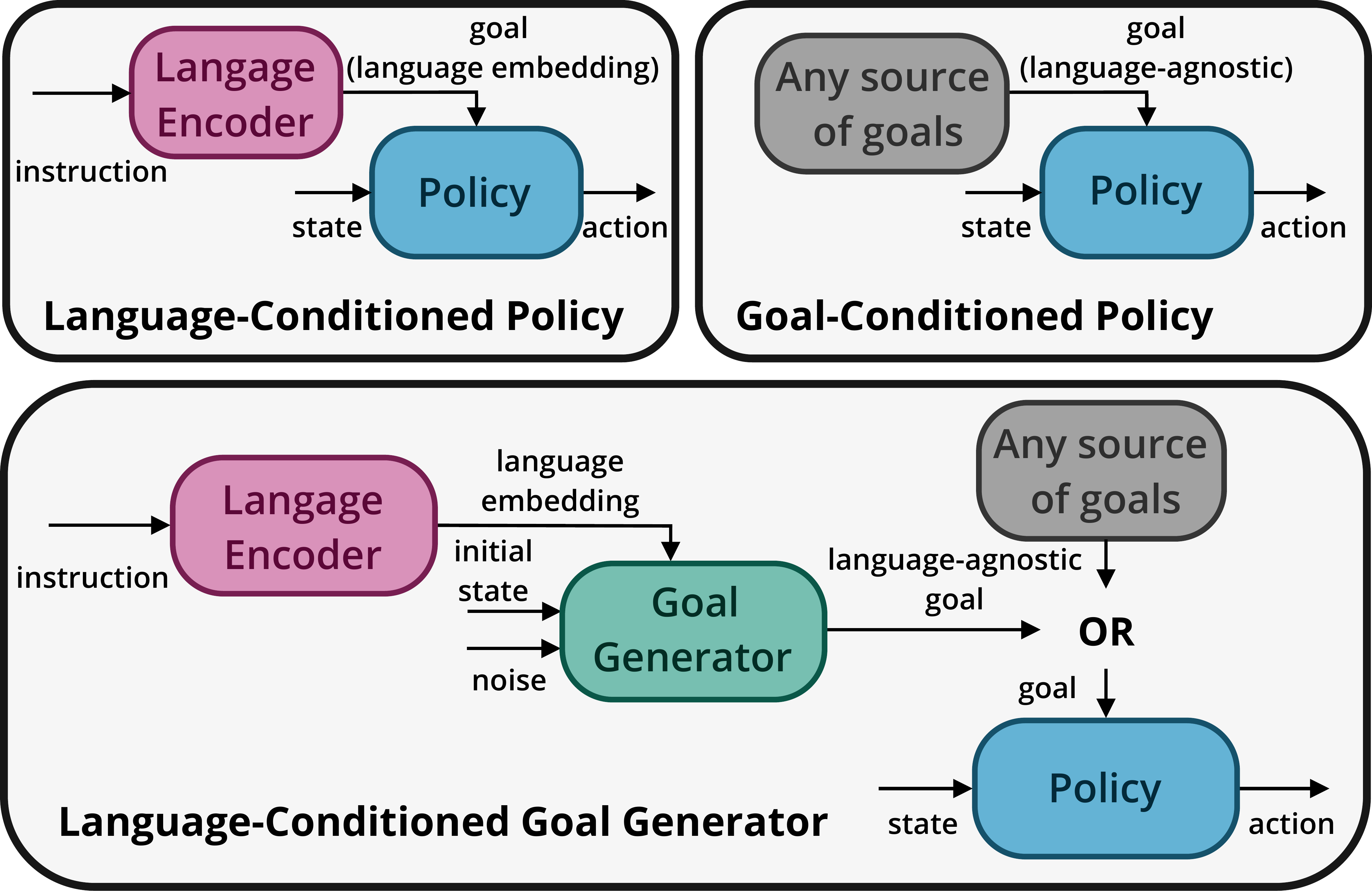}
   \caption{Language-conditioned policy, goal-conditioned policy and language-conditioned goal generator. \label{fig:schema}}
\end{figure}
This paper proposes to leverage the abstraction and generalization capacities of language while 1) decoupling language acquisition from sensorimotor learning; 2) allowing behavioral diversity. This is achieved via \textit{language-conditioned goal generation} (Figure~\ref{fig:schema}). Instead of using language to condition policies directly, we use it to condition a generator of language-agnostic goals.

Language-agnostic goals can be represented by specific states (e.g. pixels inputs) \cite{pere2018unsupervised,laversanne2018curiosity,nair2018visual,warde2018unsupervised,nair2019contextual,pong2019skew} or  handcrafted representations (e.g. target block coordinates) \cite{schaul2015universal,andrychowicz2017hindsight,florensa2018automatic,racaniere2019automated,fournier2019clic,colas2019curious}. These can come from any source: uniform sampling of the goal space \cite{schaul2015universal}, sampling from low-density areas \cite{pong2019skew}, from high learning progress areas \cite{fournier2019clic,colas2019curious}, using generative models of states \cite{nair2018visual,nair2019contextual} or generative models targeting intermediate difficulty \cite{florensa2018automatic,racaniere2019automated}. In our approach, language-conditioned goal generators are a source of language-agnostic goals among others.  As a result, pure sensorimotor training does not require any language input: agents can simply use any other source of goals.

Language-conditioned goal generators are generative models of language-agnostic goals conditioned on instructions (e.g. Variational Auto-Encoders, Generative Adversarial Networks). For any given instruction, agent can thus sample a set of matching goals. This results in behavioral diversity, a set of diverse behaviors matching any given instruction.

\paragraph{Contributions.} This paper introduces a novel approach to the problem of language grounding in \rl agents: language-conditioned goal generation. This leverages the capability of language to represent abstract goals and to  generalize, while avoiding the lack of behavioral diversity and the dependence to language inputs usually associated with language-conditioned agents. This paper presents a particular implementation of this approach and disentangles language acquisition from policy learning by assuming access to pre-trained goal-conditioned policies. Our policies are pre-trained to reach high-level configurations characterizing spatial relations between objects in the scene. 



\section{Methods}
\label{sec:methods}
This paper presents a particular implementation of  \textit{language-conditioned goal generation}. Section~\ref{sec:predicate_env} describes the learning environment and the pre-trained goal-conditioned policies. Assuming these, Section~\ref{sec:language} describes an implementation of a language-conditioned goal generator and how it is used for language grounding.

\subsection{Behavioral Policies Conditioned on Abstract Semantic Representations}
\label{sec:predicate_env}
The agents considered in this paper were pre-trained without language input by an algorithm presented in a companion paper.\footnote{This paper is not cited here to respect anonymity. It will be after reviews.} 

\begin{figure}[!ht]
  \centering
  \includegraphics[width=0.8\linewidth]{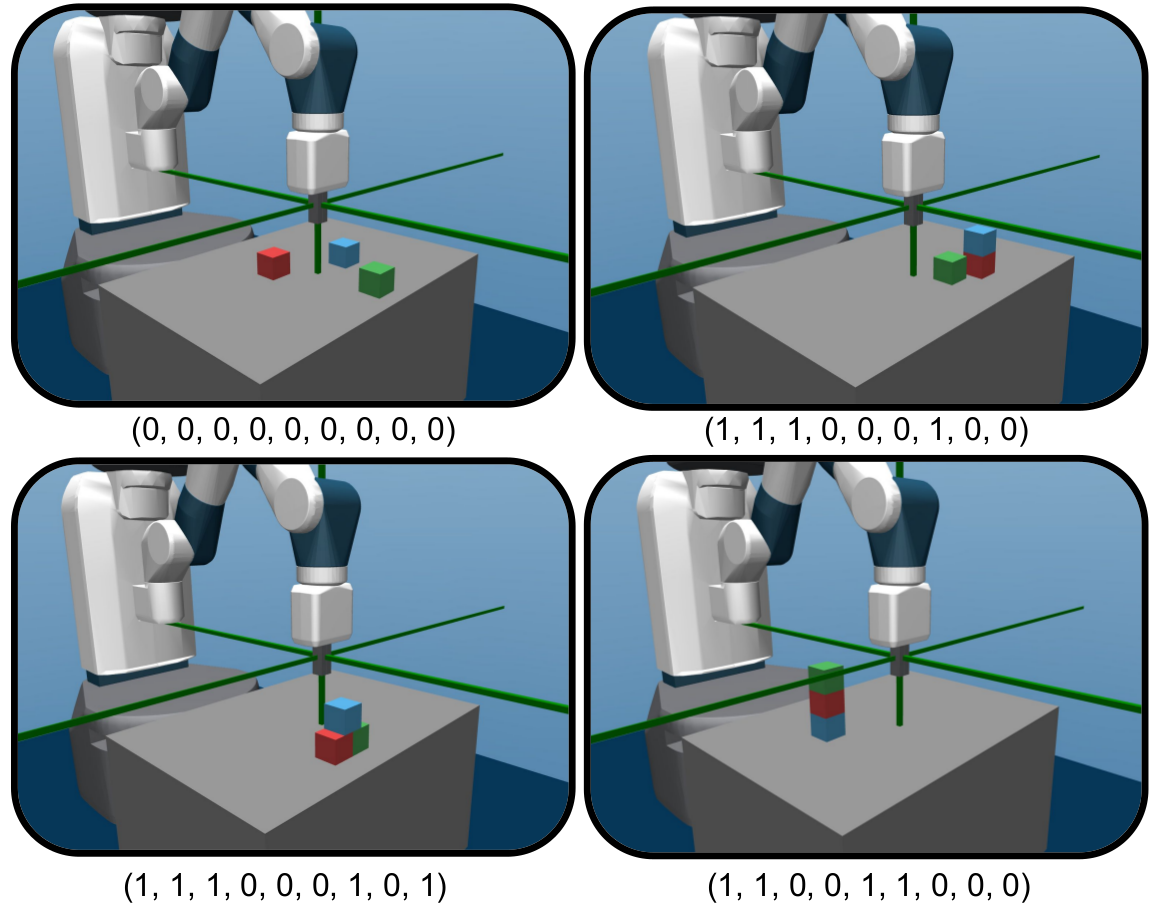}
   \caption{Some semantic configurations in \textit{Fetch Manipulate}.\label{fig:example_configs}}
\end{figure}

\paragraph{The \textit{Fetch Manipulate} environment}
Agents evolve in the \textit{Fetch Manipulate} environment: a robotic manipulation domain based on \mujoco \cite{todorov2012mujoco} and adapted from the Fetch tasks \cite{plappert2018multi}. Agents are $4$ DoF robotic arms that face $3$ colored blocks on a table (see Figure~\ref{fig:example_configs}). They are given innate representations called \textit{semantic configurations} that characterize spatial relations between blocks. During sensorimotor training, agents discovered all reachable configurations in that space and learned to master them.

\paragraph{Semantic configurations.} In contrast to traditional approaches, goals are not defined as particular targets for each block but as high-level \textit{semantic configurations}. These configurations are based on two spatial predicates infants demonstrate early in their development \cite{mandler2012spatial}: the \textit{close} and the \textit{above} binary predicates. These two predicates are applied to all permutations of object pairs, i.e. $6$ permutations for the $3$ objects we consider. Because the \textit{close} predicate is order invariant, we only need to evaluate it on $3$ object combinations. The \textit{above} predicate being order dependent, we need all $6$ permutations. The resulting binary vector of size $9$ forms the \textit{semantic configuration}. It represents the spatial relations between objects in the scene. In the resulting semantic configuration space $\{0,1\}^9$, the agent can reach $35$ physically valid configurations, including stacks of $2$ or $3$ blocks and pyramids. Supplementary Section~\ref{sec:def} provides formal definitions and properties of predicates and semantic configurations. Figure~\ref{fig:example_configs} displays visual representations of some example configurations. 

\subsection{Language-Conditioned Goal Generation}
\label{sec:language}
The language-conditioned goal generator, or \textit{language module}, generates semantic configurations matching the agent's initial configuration and a sentence describing an expected transformation of a relation between a pair of objects. This section explains its training and use for language grounding.

\paragraph{Training the language module.} The language-conditioned goal generator is implemented by a conditional Variational Auto-Encoder (\cvae) \cite{sohn2015learning} trained in a supervised setting. The training data is collected via interactions between a trained agent and a social partner. For each goal-directed trajectory the agent performs, the social partner provides the description of one of the resulting transformations in the object relations. The set of possible descriptions contains $102$ sentences, each describing in a simplified language a positive or negative shift for each of the $9$ predicates (e.g. \textit{get red above green}). \NL descriptions are encoded via a recurrent network that is jointly trained with the \cvae. Supplementary Section~\ref{sec:language_inst} provides the list of sentences and implementation details.

\paragraph{Language grounding.} At test time, agents are instructed by one of the $102$ sentences to transform a relation between objects. The trained language module acts as a translator: agent can sample a goal configuration matching their current state and instruction. This module effectively enables agents to ground \NL in their internal semantic representations and set of sensorimotor skills. We consider three evaluation settings: 1) performing a single instruction; 2) performing a sequence of instructions; 3) performing a logical combination of instructions. As the agent can generate a set of goals matching any instruction, it can easily combine these sets to perform logical functions of instructions: \textit{and} is an intersection, \textit{or} is an union and \textit{not} is the complement within the set of goals the agent discovered during sensorimotor training. Given sets of compatible goal configurations, agents can also \textit{try again}: find other goal configurations that match the required instruction when previous attempts failed.


\section{Experiments}
\label{sec:expes}

We first train a language-conditioned goal generator from a training dataset $\mathcal{D}$ collected via interactions between the agent and a social partner (Section~\ref{sec:language}). For a given initial configuration and a given sentence, we want the language module to generate all compatible final configurations, and just these.

\paragraph{Language-conditioned goal generation performance.} To evaluate the language module, we construct a synthetic, oracle dataset $\mathcal{O}$ of triplets $(c_i$, $s$, $\mathcal{C}_f(c_i,~s))$, where $c_i$ is the initial configuration, $s$ is the sentence describing the expected transformation and $\mathcal{C}_f(c_i,~s)$ is the set of all final configurations compatible with $(c_i,~s)$. Note that, on average, $\mathcal{C}_f$ in $\mathcal{O}$ contains $16.7$ configurations, while the training dataset $\mathcal{D}$ only contains $3.4$ $(20\%)$. We are interested in two metrics: 1) The \textit{Compatibility Probability} (CP) is the probability that a goal sampled from the generator belongs to $\mathcal{C}_f$; 2) The \textit{Coverage} (Cov) is the size of the intersection between $\mathcal{C}_f$ and the set resulting from sampling the language-conditioned generator $100$ times. We compute these metrics on $5$ different sets of input pairs $(c_i,s)$, each calling for a different type of generalization:

\begin{enumerate}[leftmargin=0.6cm, nolistsep]
    \item 
    Pairs found in $\mathcal{D}$, except pairs removed to form the following test sets. This calls for the extrapolation of known initialization-effect pairs $(c_i,s)$ to new final configurations $c_f$ ($\mathcal{D}$ contains only 20\% of $\mathcal{C}_f$ on average).
    \item 
    Pairs that were removed from $\mathcal{D}$, calling for a recombination of known effects $s$ on known $c_i$.
    \item 
    Pairs for which the $c_i$ was entirely removed from $\mathcal{D}$. This calls for the transfer of known effects $s$ on unknown $c_i$.
    \item 
    Pairs for which the $s$ was entirely removed from $\mathcal{D}$. This calls for generalization in the language space, to generalize unknown effects $s$ from related sentences and transpose this to known $c_i$.
    \item 
    Pairs for which both the $c_i$ and the $s$ were entirely removed from $\mathcal{D}$. This calls for the generalizations 3 and 4 combined.
\end{enumerate}
Our language module demonstrates these $5$ types of generalization (see Table~\ref{tab:lang}). Agents can generate goals from situations they never encountered (Test 3). They can generalize the meaning of sentences they never heard (Test 4) and even apply the latter to unknown situations (Test 5). 
We detail the testing sets in Supplementary Section~\ref{sec:language_inst}.

\begin{table}[ht!]
    \caption{Language module average metrics over $10$ seeds. Std is below $0.07$ for Cov and $0.06$ for CP.\label{tab:lang}}
    \centering
    \vspace{0.2cm}
    \begin{tabular}{l|ccccc}
        Metr. & Test 1 & Test 2 & Test 3 & Test 4 & Test 5 \\ 
        \hline
        CP & 0.93 & 0.94 &  0.95 & 0.90 & 0.92 \\
        Cov & 0.97 & 0.93 &  0.98 & 0.99 & 0.98 
    \end{tabular}
\end{table}

\paragraph{Grounding language in sensorimotor behavior.} We investigate how the language module interacts with the sensorimotor skills of the agent. We consider three evaluation settings. In the \textit{transition} setup, we look at the average success rate of the agent when asked to perform the $102$ instructions $5$ times each, resetting the environment each time. In the \textit{expression} setup, we evaluate the agent on $500$ randomly generated logical functions of sentences. In both setups, we give the agent $5$ attempts, enabling it to resample new compatible goals when the previous failed (without reset). Success rates after $1$ (SR1) and $5$ (SR5) attempts are reported in Table~\ref{tab:grounding}. In the \textit{sequence} setup, we ask the agent to execute $20$ random sequences of instructions without reset and report the average number of successes before the agent fails: $N_s~=~14.9\pm5.7$ (std). The $10$ \rl agents are evaluated with the $10$ \textsc{c-vae} models evaluated above. These results show that the language module efficiently implements language grounding. Agents achieve instructed transitions almost all the time, resampling alternative goals when previous ones failed. They only fail when a previous trajectory kicked the blocks out of reach.

 \begin{table}[h!]
    \caption{Language grounding performance metrics over 10 seeds (av$\pm$std). \label{tab:grounding}}
    \centering
    \vspace{0.2cm}
    \begin{tabular}{l|cccc}\
        Metr. & Transition & Expression \\ 
        \hline
        SR1 & $0.89\pm0.05$ & $0.74\pm0.08$ \\  
        SR5 & $0.99\pm0.01$ & $0.94\pm0.06$
    \end{tabular}
\end{table}

\section{Discussion}
This paper introduces \textit{language-conditioned goal generation}: a new approach to the problem of language grounding in \rl agents. It shows the following advantages over traditional language-conditioned \rl agents:

\begin{itemize}[leftmargin=0.6cm, nolistsep]
    \item 
    \textbf{Decoupled language grounding.} Our approach enables agents to decouple language acquisition from sensorimotor learning, as it is observed in infants \cite{wood1976role}. However, nothing in the architecture prevents language from being grounded during sensorimotor learning. This would result in "overlapping waves" of sensorimotor and linguistic development \cite{siegler1998emerging}. 
    \item 
    \textbf{Behavioral diversity.} The language module generates a diversity of goals matching any instruction (see Coverage metrics in Table~\ref{tab:lang}). This results in a behavioral diversity that language-conditioned agents cannot demonstrate. Indeed, while a language-conditioned agent trained on \textit{put red close\_to green} would only push the red block towards the green one, our agent can generate many matching goal configurations. It could build a pyramid, make a blue-green-red pile or target a dozen other compatible configurations. 
    \item 
    \textbf{Trying again.} Generating a diversity of goals matching any instruction enables agents to \textit{try again}: to find alternative approaches to satisfy a same instruction when previous attempts failed. Table~\ref{tab:grounding} shows that benefiting from several attempts significantly improves the chances of success. This could also improve robustness to perturbed environments where former successful behaviors fail.
    \item 
    \textbf{Logical expressions.} Generating sets of compatible goals makes it easy to scale language understanding to any logical combination of instructions. This cannot be achieved with language-conditioned policies.
\end{itemize}

One strength of the language-conditioned approach is their capability of \textit{systematic generalization} \cite{hermann2017grounded,bahdanau2018learning,hill2019emergent,colas2020language}. Our language module also seems to demonstrate generalization abilities: agents can generate goals from situations they never encountered (Type 3). They can generalize the meaning of sentences they never heard (Type 4) and even apply the latter to unknown situations (Type 5). Type 4, especially, involves recombinations of action verbs and attributes similar to \citet{hill2019emergent, colas2020language}.

\cvae were already used to generate goals matching initial states in \citet{nair2019contextual}. However, the additional condition on instructions brings the benefits of language use: 1) the representation of abstract goals; 2) the systematic generalization capacities. In addition instructions enable agents to control their goal generation. Because image-based goal generation was shown to work in \citet{nair2018visual,pong2019skew,nair2019contextual}, we believe our language-conditioned goal generator could be trained to generate imaged-based goals.

Further work could investigate the extension of language-conditioned goal generator to diverse goal-conditioned settings such as, for example, \textit{Quality-Diversity} algorithms. These algorithms train populations of diverse and high-performing solutions to a problem \cite{cully_qd}. Each solution is associated with its \textit{behavioral characterization}: a low-dimensional description of its behavior. Our language-conditioned goal generator could be used to map language instructions to that behavioral space, enabling a language-based control of these diversity-seeking evolutionary algorithms \cite{mapelite,nslc,nses,colas2020scaling}.

\section*{Acknowledgments}
C\'edric Colas is partly funded by the French Minist\`ere des Arm\'ees - Direction G\'en\'erale de l’Armement.

\small
\bibliographystyle{icml2020}
\bibliography{references}

\clearpage
\section*{Supplementary Material}
This supplementary material includes:
\begin{itemize}[leftmargin=0.6cm, nolistsep]
    \item Section~\ref{sec:def}: a formal definition of semantic configurations and the definiton of the semantic configurations used in the \textit{Fetch Manipulate} environment.
    \item Section~\ref{sec:language_inst}: further details about the training of our language-conditioned goal generation module.
\end{itemize} 

\section{Formal Definition of Semantic Configurations}
\label{sec:def}

Semantic configurations are based on a collection of formal systems known as \textit{predicate logic} or \textit{first-order logic}. They use \textit{$k$-ary relations} to describe possible connections between $k$ quantified variables. This paper focuses on \textit{spatial binary predicates} characterizing spatial relations between pairs of physical objects. We provide formal definitions, properties and examples below.

\paragraph{Binary predicates} Consider a finite set of objects $O~=~\{o_1,~o_2,~...,~o_M\}$. A binary predicate $p$ associated with a semantic relation \textbf{r} is an expression that takes as input any ordered pair of objects $(o_i,~o_j) \in O^2$. $p(o_i,~o_j)$ is said \textit{true} if and only if "$o_i$ \textbf{r} $o_j$" is verified. For simplicity, we refer to $p$ and \textbf{r} interchangeably. 

\paragraph{Examples of binary predicates.} We consider the objects $o_1$ and $o_2$.
\begin{itemize}
    \item The expression "$o_1$ is \textbf{close} to $o_2$" describes the predicate \textit{close} evaluated on ($o_1$, $o_2$).
    \item The expression "$o_2$ is \textbf{above} $o_1$" describes the predicate \textit{above} evaluated on ($o_2$, $o_1$).
\end{itemize}

\paragraph{Semantic mapping functions.}
To achieve symbol grounding into non-symbolic sensorimotor interactions using predicates, we define a \textit{semantic mapping function} $f$ associated with the binary predicate $p$ as the probability that $p$ is true given the states of the considered objects. Formally, if we consider the objects $o_i$, $o_j$ and their respective states $s_i$, $s_j$, then:
\begin{align*}
    f(s_i,~s_j)~=~\textrm{P }(p(o_i,~~o_j)~|~s_i,~s_j).
\end{align*}
This paper assumes oracle deterministic semantic mapping functions, i.e. $f$ is a Boolean function in $\{0,1\}$. Practically, we hard-code a function, assumed internal to the agent, that uses predefined fixed thresholds to determine whether a predicate is true or false given the states of the considered objects. For example, for the \textit{close} predicate, it outputs $1$ if and only if the Euclidean distance between the two considered objects is below a defined threshold. For the sake of simplicity, we omit the word \textit{deterministic}.

\paragraph{Symmetry and asymmetry.}
Consider a finite set of objects $O~=~\{o_1,~o_2,~...,~o_M\}$ and a binary predicate $p$. The predicate $p$ is said to be \textit{symmetric} if and only if, for any ordered pair of objects $(o_i,~o_j)~ \in~ O^2$, "$o_i$ \textbf{r} $o_j$" and "$o_j$ \textbf{r} $o_i$" are equivalent. As a result, the corresponding semantic mapping function $f$ needs to be symmetric, i.e. $f(o_i,~o_j)~=~f(o_j,~o_i)$. The predicate $p$ is said to be \textit{asymmetric} iff, for any ordered pair $(o_i,~o_j)~ \in~ O^2$, "$o_i$ \textbf{r} $o_j$" implies \textbf{not} "$o_j$ \textbf{r} $o_i$".

\paragraph{Examples. } We consider the objects $o_1$ and $o_2$.
\begin{itemize}
    \item \textit{close} is symmetric: "$o_1$ is \textbf{close} to $o_2$" $\Leftrightarrow$ "$o_2$ is \textbf{close} to $o_1$". The corresponding semantic mapping function is based on the Euclidean distance, which is symmetric.
    \item \textit{above} is asymmetric: "$o_1$ is \textbf{above} $o_2$" $\Rightarrow$ \textbf{not} "$o_2$ is \textbf{above} $o_1$". The corresponding semantic mapping function evaluates the sign of the difference of the object $Z$-axis coordinates.
\end{itemize}

\paragraph{Effective number of predicate relations.} Consider a finite set of $M$ objects $O~=~\{o_1, o_2, ..., o_M\}$ and a binary predicate $p$.
\begin{itemize}
    \item 
    If $p$ is not symmetric, then the effective number of relations $K_p$ that can be described without redundancy is equal to the number of \textbf{permutations} of $2$ objects among $M$, i.e. $K_p~=~A_{M,2}~=~M(M-1)$.
    \item 
    If $p$ is symmetric, then the effective number of relations $K_p$ is equal to the number of \textbf{combinations} of $2$ objects among $M$, i.e. $K_p~=~{M\choose 2}~=~\frac{M(M-1)}{2}$.
\end{itemize}

\paragraph{Semantic configurations based on spatial relations.}
Let $(p_i)_{i\in[1..P]}$ be a list of $P$ binary predicates. The concatenation of the evaluations of the semantic mapping functions $f_i$ on the $K_{pi}$ pairs of objects forms a \textit{semantic configuration}. It is an abstract representation of a scene which characterizes all relations defined by the $(p_i)$ predicates among the $M$ objects. This defines a binary \textit{semantic configuration space} $\mathcal{C}_p~=~\{0,1\}^{K_c}$, where $K_c~=~\sum^P_{i=1}K_{p_i}$. If any world configuration can be mapped to $\mathcal{C}_p$, not all configurations are reachable (e.g. $o_1$ cannot be \textit{above} and \textit{below} $o_2$ at the same time).

\paragraph{Semantic representation space in \textit{Fetch Manipulate}.}
In the \textit{Fetch Manipulate} environment, we restrict semantic representations to the use of the \textit{close} and \textit{above} binary predicates applied on $M=3$ objects. The resulting semantic configurations are formed by: 
\begin{align*}
    c_p~=&~[\textit{c}(o_1,o_2),~\textit{c}(o_1,o_3),~\textit{c}(o_2,o_3),~ \textit{a}(o_1,o_2),\\
    &\textit{a}(o_2,o_1),~\textit{a}(o_1,o_3),~\textit{a}(o_3,o_1),~\textit{a}(o_2,o_3),~\textit{a}(o_3,o_2)],
\end{align*}

where \textit{c()} and \textit{a()} refer to the \textit{close} and \textit{above} predicates respectively and $(o_1,~o_2,~o_3)$ are the red, green and blue blocks respectively.


\section{Language-Conditioned Goal Generator Training}
\label{sec:language_inst}
We use a conditional Variational Auto-Encoder (\textsc{c-vae}) \cite{sohn2015learning}. Conditioned on the initial configuration and a sentence describing the expected transformation of one object relation, it generates compatible goal configurations. After the first phase of goal-directed sensorimotor training, the agent interacts with a hard-coded social partner as described in Main Section~\ref{sec:language}. From these interactions, we obtain a dataset of $5000$ triplets: initial configuration, final configuration and sentence describing one change of predicate from the initial to the final configuration. The list of sentences used by the synthetic social partner are provided in Table~\ref{tab:list_inst}. Note that \textit{red}, \textit{green} and \textit{blue} refer to objects $o_1,~o_2,~o_3$ respectively.

\paragraph{Content of test sets.}We describe the $5$ test sets:
\begin{enumerate}
    \item 
    Test set $1$ is made of input pairs $(c_i,~s)$ from the training set, but tests the coverage of all compatible final configurations $\mathcal{C}_f$, 80\% of which are not found in the training set. In that sense, it is partly a test set.
    \item 
    Test set $2$ contains two input pairs: \{$[0~1~0~0~0~0~0~0~0]$, \textit{put blue close\_to green}\} and \{$[0~0~1~0~0~0~0~0~0]$, \textit{put green below red}\} corresponding to $7$ and $24$ compatible final configurations respectively.
    \item 
    Test set $3$ corresponds to all pairs including the initial configuration $c_i~=~[1~1~0~0~0~0~0~0~0]$ ($29$ pairs), with an average of $13$ compatible final configurations.
    \item 
    Test set $4$ corresponds to all pairs including one of the sentences \textit{put green on\_top\_of red} and \textit{put blue far\_from red}, i.e. $20$ pairs with an average of $9.5$ compatible final configurations.
   \item 
    Test set $5$ is all pairs that include both the initial configuration of test set $3$ and one of the sentences of test set $4$, i.e. 2 pairs with $6$ and $13$ compatible goals respectively. Note that pairs of set $5$ are removed from sets $3$ and $4$.
\end{enumerate}

\begin{table}[!h]
    \centering
    \caption{List of instructions. Each of them specifies a shift of one predicate, either from false to true ($0\to1$) or true to false ($1\to0$). \textbf{block A} and \textbf{block B} represent two different blocks from \{red, blue, green\}.
    \label{tab:list_inst}}
    \vspace{0.5cm}
    \begin{tabular}{l|l}
      Type & Sentences \\
     \hline
  Close & \textit{Put \textbf{block A} close\_to \textbf{block          B}}, \\
$0\to1$ & \textit{Bring \textbf{block B} and \textbf{block A}             together},\\
        
        & \textit{Put \textbf{block B} close\_to \textbf{block A}}, \\
        & \textit{Bring \textbf{block A} and \textbf{block B} together},\\ 
        & \textit{Get \textbf{block B} and \textbf{block A} close\_from each\_other},\\
        & \textit{Get \textbf{block A} close\_to \textbf{block B}}\\
        & \textit{Get \textbf{block A} and \textbf{block B} close\_from each\_other}, \\
        & \textit{Get \textbf{block B} close\_to \textbf{block A}}.\\
        \hline
  Close & \textit{Put \textbf{block A} far\_from \textbf{block          B}}, \\
$1\to0$ & \textit{Get \textbf{block A} far\_from \textbf{block            B}}, \\ 
        & \textit{Put \textbf{block B} far\_from \textbf{block A}}, \\
        & \textit{Get \textbf{block B} far\_from \textbf{block A}}, \\ 
        & \textit{Get \textbf{block A} and \textbf{block B} far\_from each\_other}, \\
        & \textit{Bring \textbf{block A} and \textbf{block B} apart}, \\ 
        & \textit{Get \textbf{block B} and \textbf{block A} far\_from each\_other}, \\
        & \textit{Bring \textbf{block B} and \textbf{block A} apart}. \\ 
        \hline

Above 
        & \textit{Put \textbf{block A} above \textbf{block B}}, \\
$1\to0$ & \textit{Put \textbf{block A} on\_top\_of \textbf{block          B}}, \\ 
        & \textit{Put \textbf{block B} under \textbf{block A}}, \\
        & \textit{Put \textbf{block B} below \textbf{block A}}. \\ 
    
    \hline

Above 
        & \textit{Remove \textbf{block A} from\_above \textbf{block B}}, \\
$1\to0$ & \textit{Remove \textbf{block A} from \textbf{block              B}},\\ 
        & \textit{Remove \textbf{block B} from\_below \textbf{block A}},\\
        &\textit{Put \textbf{block B} and \textbf{block A} on\_the\_same\_plane}, \\ 
        & \textit{Put \textbf{block A} and \textbf{block B} on\_the\_same\_plane}. \\ 
      \end{tabular}

\end{table}

\newpage
\paragraph{Testing on logical expressions of instructions.} To evaluate our agents on logical functions of instructions, we generate three types of expressions:
\begin{enumerate}
    \item 
    $100$ instructions of the form "A and B" where A and B are basic instructions corresponding to shifts of the form \textit{above}~$0~\to~1$ (see Table~\ref{tab:list_inst}). These intersections correspond to stacks of $3$ or pyramids.
    \item 
    $200$ instructions of the form "A and B" where A and B are \textit{above} and \textit{close} instructions respectively. B can be replaced by "not B" with probability 0.5.
    \item 
    $200$ instructions of the form "(A and B) or (C and D))", where A, B, C, D are basic instructions: A and C are \textit{above} instructions while B and D are \textit{close} instructions. Here also, any instruction can be replaced by its negation with probability 0.5. 
\end{enumerate}

\paragraph{Hyperparameters.} The encoder is a fully-connected neural network with two layers of size $128$ and \textit{ReLU} activations. It takes as input the concatenation of the final binary configuration and its two conditions: the initial binary configuration and an embedding of the \NL sentence. The \NL sentence is embedded with an recurrent network with embedding size $100$, \textit{tanh} non-linearities and biases. The encoder outputs the mean and log-variance of the latent distribution of size $27$. The decoder is also a fully-connected network with two hidden layers of size $128$ and \textit{ReLU} activations. It takes as input the latent code $z$ and the same conditions as the encoder. As it generates binary vectors, the last layer uses \textit{sigmoid} activations. We train the architecture with a mixture of Kullback-Leibler divergence loss $(KD_{\text{loss}})$ w.r.t a standard Gaussian prior and a binary Cross-Entropy loss $(BCE_{\text{loss}})$. The combined loss is $\text{loss}~=~BCE_{\text{loss}}~+~\beta~\times~KD_{\text{loss}}$ with $\beta~=~0.6$. We use an Adam optimizer with a learning rate of $5\times10^{-4}$, a batch size of $128$ and optimize for $150$ epochs. As training is fast ($\approx2$ min on a single cpu), we conducted a quick hyperparameter search over $\beta$, layer sizes, learning rates and latent sizes (see Table~\ref{tab:hyper}). We found robust results for various layer sizes, various $\beta$ below $1$ and latent sizes above $9$.

\begin{table}[h!]
    \centering
    \caption{Language module hyperparameter search. In bold are the selected hyperparameters.\label{tab:hyper}}
    \vspace{0.2cm}
    \begin{tabular}{l|c}
        Hyperparam. &  Values. \\
        \hline
        $\beta$ & $[0.5,~\textbf{0.6},~0.7,~0.8,~0.9,~1.]$ \\
        layers size & $[\textbf{128},~256]$ \\
        learning rate & $[0.01,~\textbf{0.005},~0.001]$ \\
        latent sizes & $[9,~18,~\textbf{27}]$ \\
    \end{tabular}
\end{table}

\end{document}